\documentclass{article}

\usepackage[preprint]{cpal_2026}

\usepackage{graphicx}
\usepackage{wrapfig}
\usepackage{url}
\usepackage{hyperref}
\usepackage{times}  
\usepackage{helvet}  
\usepackage{courier}  
\usepackage{graphicx} 
\usepackage{natbib}  
\usepackage{caption} 
\usepackage{enumitem} 
\usepackage{amssymb}

\usepackage{amsmath}
\usepackage{algorithm}
\usepackage{algorithmic}
\usepackage{booktabs}    
\usepackage{caption}     
\usepackage{array}       
\usepackage{xcolor}
\usepackage{colortbl}
\usepackage{booktabs}

\usepackage{newfloat}
\usepackage{listings}

\title{Data-Efficient and Robust Trajectory Generation through  Pathlet Dictionary Learning}

\author{%
  Yuanbo Tang\textsuperscript{1}, ~Yan Tang\textsuperscript{2}, ~Zixuan Zhang\textsuperscript{1}, ~Zihui Zhao\textsuperscript{1}, ~Yang Li\textsuperscript{1}\thanks{Corresponding author.} \\
  \textsuperscript{1}Tsinghua University International Campus Phase I, Nanshan District, Shenzhen, \textsuperscript{2}College of Software, Northeastern University, Shenyang, Liaoning Province\\
  \texttt{tori2011@gmail.com}
}

\begin{document}

\maketitle

\begin{abstract}

Trajectory generation has recently drawn growing interest in privacy-preserving urban mobility studies and location-based service applications. 
Although many studies have used deep learning or generative AI methods to model trajectories and achieved promising results, real‑world trajectory data are noisy and often incomplete (e.g., device instability, low sampling rates, privacy‑driven partial reporting), introducing distribution shifts and, as observed in our experiments, marked differences between synthetic and real trajectory distributions.
To address this issue, we exploit the low-dimensional structure and regular patterns in 
urban trajectories and propose a parsimonious deep generative model based on sparse pathlet 
representations, which encode trajectories with sparse binary vectors associated with a 
learned compact dictionary of trajectory segments.
Specifically, we introduce a probabilistic graphical model to describe the trajectory generation process, which includes a Variational Autoencoder (VAE) component and a linear decoder component. 
During training, the model can simultaneously learn the latent embedding of sparse pathlet representations and the pathlet dictionary that captures essential mobility patterns in the trajectory dataset. The conditional version of our model can also be used to generate customized trajectories based on temporal and spatial constraints.
Our model can effectively learn data distribution even using noisy data, achieving relative improvements of $35.4\%$ and $26.3\%$ over strong baselines on two real-world trajectory datasets.
Moreover, the generated trajectories can be conveniently utilized for multiple downstream tasks, including trajectory prediction and data denoising.
Lastly, the framework design offers a significant efficiency advantage, saving $64.8\%$ of the time and $56.5\%$ of GPU memory compared to previous approaches. 
The code repository is available at this \href{https://anonymous.4open.science/r/Data-Efficient-and-Robust-Trajectory-Generation-through-Pathlet-Dictionary-Learning-045E}{URL}.
\end{abstract}

\section{Introduction}

GPS trajectories contain a wealth of useful information that can be utilized for tasks such as urban planning, traffic management, and location-based services. 
However, this field also faces several challenges. Firstly, there are significant concerns about privacy leaks. Secondly, in scenarios like newly developed or developing regions, high-quality GPS trajectory data is often scarce due to limited infrastructure.
These factors have hindered the development and large-scale application of trajectory-related mining algorithms.

Generative models are a promising solution to these challenges.
By synthesizing new datasets that follow the same distribution as the original data, trajectory generative models act as privacy-preserving surrogates, enabling advanced analytical tasks without compromising sensitive user information.
Besides, customized trajectory generation is also useful for downstream tasks such as navigation and travel route planning.
Therefore, this field has garnered widespread interest from researchers in recent years. For instance, Wang  et al. proposed a deep learning model called MTNet \cite{wangDeepGenerativeModel2022}, which considers the topological structure of road networks and road knowledge.  Similarly, \cite{shiGRAPHCONSTRAINEDDIFFUSIONENDTOEND2024} proposed a model named GDP, which is a diffusion-based model for end-to-end data-driven path planning and trajectory generation that outperforms previous frameworks by incorporating user intentions and road network constraints. However, these deep learning-based methods are typically data-hungry \cite{lecun2015deep} and prone to overfitting when training data is limited.

\begin{figure}[t]
\centering
\includegraphics[width=0.88\textwidth]
{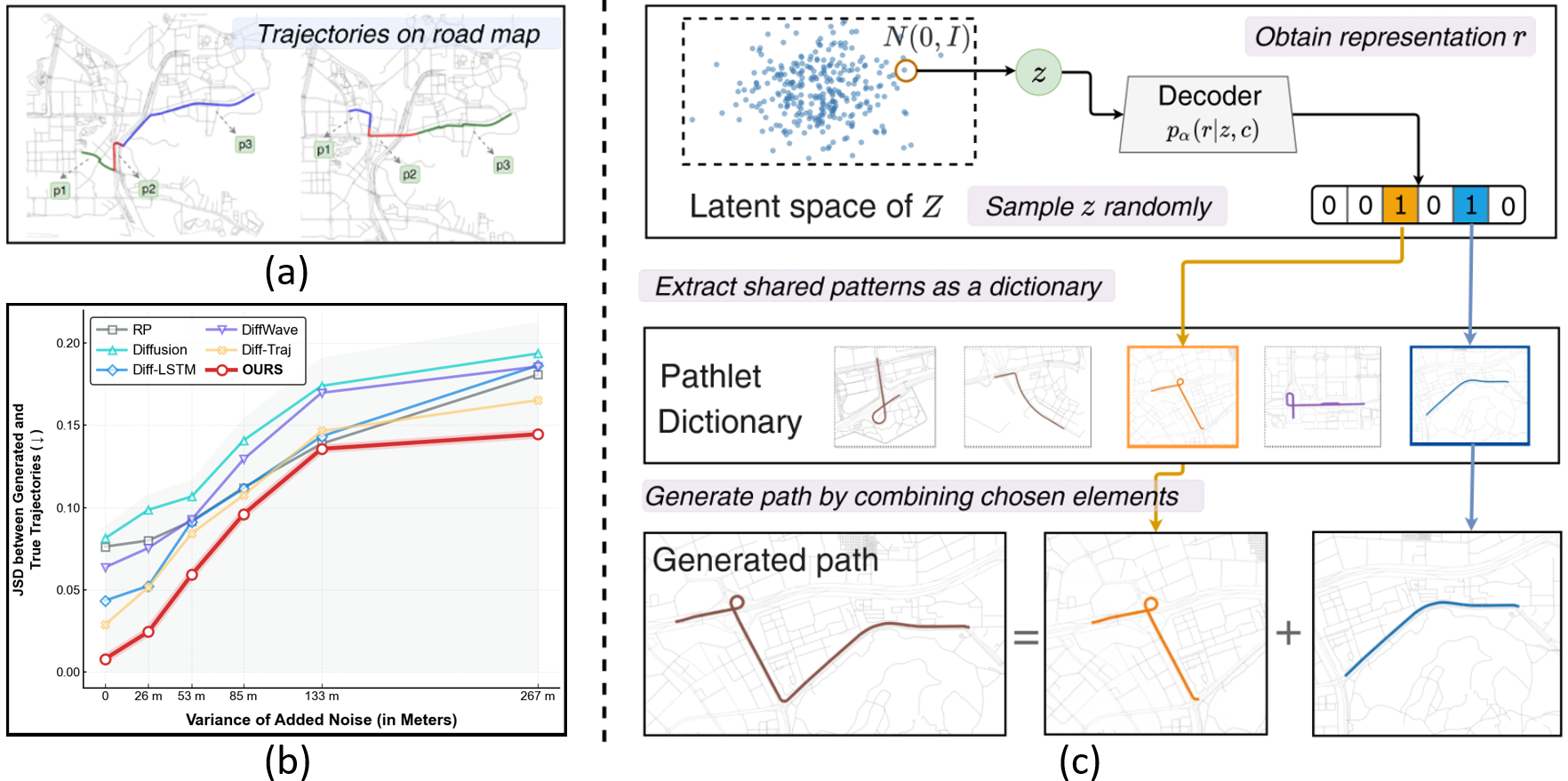}
\captionsetup{skip=5pt}
\caption{\label{fig:pic1}  (a) Example of pathlet composition; (b) Comparison of the impact of noise on the distributional fidelity (measured by JSD) of different methods, showing the consistent robustness of our approach; (c) Schematic overview of the framework, supporting both road network-based and grid-based trajectory representations.}
\end{figure}

Another challenge for existing trajectory generative models is their reliance on high-quality training data. In real-world deployments, trajectories are often noisy and incomplete due to device instability, environmental interference, low sampling rates, and privacy-driven partial reporting \cite{liuLightTRLightweightFramework2024,liKnowledgebasedTrajectoryCompletion2016}. 
These artifacts induce distributional shift and systematic bias; moreover, empirical evidence from  our experiments shows that deep generative models degrade markedly when trained on such corrupted inputs (see Figure~\ref{fig:pic1} (b)).
Although preprocessing techniques—such as interpolation, filtering, and multi-source data fusion—can partially alleviate these issues before model training \cite{renMTrajRecMapConstrainedTrajectory2021,yinyifangFeaturebasedMapMatching2018}, they incur additional computational cost and can introduce errors that propagate to the generation stage.



To address these challenges, we propose a \textbf{robust} trajectory generation framework, as illustrated in Figure \ref{fig:pic1}. This approach is based on the empirical finding that urban trajectories exhibit strong spatiotemporal regularities \cite{gonzalez2008understanding}.
Spatial regularity at a given time can be seen as frequently traveled sub-paths, also known as \textit{pathlets} in \cite{chenPathletLearningCompressing2013b,tangExplainableTrajectoryRepresentation2023}. Pathlets serve as the basic units of a trajectory, similar to how words are the basic units of a corpus. Each trajectory can be represented by a binary vector indicating the pathlet index.
Given a dictionary of pathlets learned from a large collection of trajectories, we can generate new trajectories by selecting and concatenating elements from the dictionary. 

Specifically, we design a probabilistic graphical model to describe the trajectory generation process, implemented using a Variational Autoencoder (VAE) component and a linear dictionary decoder component. The VAE is utilized to model the distribution of binary representation vector, while the linear decoder converts the representation vectors to trajectories.
During the training process, pathlet dictionary and the VAE model for representation vectors are jointly learned. The pathlet learning component aims to find a compact dictionary that reconstructs each trajectory with the least number of elements, resulting in a sparse representation. 
This parsimonious design enhances robustness and strengthens the ability to retrieve the true underlying signals from noisy data.
Moreover, elements in the dictionary can be visualized as important route segments on the road map, making the model's behavior  interpretable.

To our knowledge, this work is the first to introduce dictionary learning and    compositional learning approaches in trajectory generative models.
We validated our approach on two real-world datasets and the experiments revealed that our method generates data more similar to the real data compared to other methods under different levels of noise. Additionally, we further verify the effectiveness of our method in downstream tasks including data recovery and conditional generation.

Overall, our main contributions can be summarized as follows:
\begin{itemize}[leftmargin=*,itemindent=0pt]
    \item We propose a data-efficient and robust trajectory generation framework that integrates the classic VAE model with the compositional representation provided by pathlet dictionary learning. Bernoulli sampling is introduced to the VAE reparameterization step to generate binary data.
    
    \item We thoroughly evaluated the performance of our method on real-world datasets, and the results show that our method surpasses previous approaches in terms of the Jensen-Shannon Divergence (JSD) score. This validates that our model is more effective in modeling trajectory data distributions in the presence of noise and data scarcity.
    \item We further validated the utility of the algorithm on two downstream tasks: trajectory denoising and conditional generation. Additionally, we visualized key results and discussed the interpretability of the method. We further analyzed the efficiency of the proposed method.

\end{itemize}

\section{Related work}

\subsection{Trajectory generative model.} 
Existing trajectory generation models can be divided into traditional methods and deep learning-based methods. Traditional methods include combining trajectories from the dataset or perturbing the original trajectories to generate new ones \cite{nergizTrajectoryAnonymizationGeneralizationbased2008,zandbergenEnsuringConfidentialityGeocoded2014}. These approaches rely on ad-hoc design, often resulting in datasets with significant distributional differences from the original data, leading to lower usability.

In an effort to solve these problems, recent trends in this field have used deep neural networks to model the intrinsic spatio-temporal distribution of trajectories, with subsequent data generation by sampling from the learned distribution.
Researchers formulate raw trajectory data in various formats as inputs for neural networks. Some methods segment GPS sequences into grids, allowing neural networks to learn distribution patterns between grids \cite{heDPTDifferentiallyPrivate2015,kongMobilityDatasetGeneration2018}. However, due to the variability in data distribution across different regions, determining the appropriate grid scale often proves to be a challenging task. Other approaches involve mapping raw GPS data onto road networks and then learning the distribution of edge sequences \cite{choiTrajGAILGeneratingUrban2021,wuModelingTrajectoriesRecurrent2017a}. 
The effectiveness of these schemes has been validated in data generation as well as in multiple downstream tasks.
Our work, on the other hand, focuses not only on effectiveness but also on robustness and interpretability. Moreover, our framework is flexible enough to support both grid-based and road network-based trajectory representations, making it applicable to trajectory datasets generated in road network environments as well as those collected in free-space scenarios.



\subsection{Dictionary learning and sparse recovery.}
Our framework design incorporates the concepts of dictionary learning and sparse recovery. Dictionary learning focuses on the training stage with the goal of learning a set of patterns from the dataset. On the other hand, sparse recovery focuses on the application stage, using the learned dictionary to sparsely represent data. 
Dictionary learning and sparse recovery  have been widely applied in various fields \cite{yangImageSuperresolutionSparse2010a,wrightRobustFaceRecognition2009,mairalSparseRepresentationColor2008}.
Recently, some studies have combined deep neural networks and dictionary learning to leverage the strengths of both. For example, \cite{zheng_deep_2021} proposed deep convolutional dictionary learning (DCDicL) framework, which adaptively adjusts dictionaries for each image to enhance image restoration, particularly improving denoising performance by preserving subtle structures and textures. 

\subsection{Dictionary learning for trajectories.}
Early work on dictionary learning for trajectory data includes the introduction of the pathlet concept, which was initially proposed by \cite{chenPathletLearningCompressing2013b} and further developed in subsequent papers \cite{alixPathletRLTrajectoryPathlet2023}. For example, \cite{tangExplainableTrajectoryRepresentation2023} used a randomized rounding algorithm to further optimize the dictionary and explored the effectiveness of trajectory representation across multiple downstream tasks. Previous methods used selection-based strategies for dictionary learning, which were time-consuming. Our approach, based on learning rather than selection strategies, significantly improves efficiency.

\section{Preliminary}

\noindent\textbf{Terminology.}
Given a trajectory dataset $X$, we consider two representation scenarios: (1) \textit{Road network mapping}: trajectories are mapped onto a road network that can be formed as a directed graph $G=(E,V)$, where a trajectory $\boldsymbol{t}\in X$ is defined as a sequence of edges $e$ on $G$, denoted by ${\boldsymbol{t}}=\{e_1,e_2,...,e_n\}$; (2) \textit{Grid-based mapping}: trajectories are represented in a discretized spatial grid where each trajectory $\boldsymbol{t}\in X$ is defined as a sequence of grid cells $c$, denoted by ${\boldsymbol{t}}=\{c_1,c_2,...,c_n\}$. 
Similarly, a pathlet $p$ refers to a sequence of edges (in road network scenario) or grid cells (in grid-based scenario), and all pathlets together form a dictionary.

\noindent\textbf{Trajectory Vectorization.} Each variable-length trajectory $\boldsymbol{t}$ can be transformed into a binary vector $\boldsymbol{x}$ with the size of $|E| \times 1$, where $|E|$ refers to the number of spatial units (edges in road network mapping or grid cells in grid-based mapping). Each entry $\boldsymbol{x}_{i}=1$ if the corresponding spatial unit is covered by this trajectory, and $\boldsymbol{x}_{i}=0$ otherwise. Similarly, the binary vector $\boldsymbol{r}$ is used to store which elements from the dictionary $D$ are selected for reconstruction.

\noindent\textbf{Problem statement (Trajectory Generation).}
Given a trajectory dataset $X$ represented either through road network mapping or grid-based discretization, the goal of the trajectory generation task is to learn a generative model $M$ using  $X$, which can synthesize trajectory data $Y$ that satisfies the following:
\begin{itemize}[leftmargin=*,itemindent=0pt]
\item {\em Authenticity}: The generated trajectories $Y$ retain the spatiotemporal properties and distribution of the original trajectory dataset $X$. 
\item {\em Utility}: The generated trajectories and the generative  model $M$ can benefit various downstream tasks.
\end{itemize}
Besides the above two criteria that are common to any generative models, an ideal  model $M$ should also have:
\begin{itemize}[leftmargin=*,itemindent=0pt]
\item {\em Robustness and Interpretability}: The algorithm can effectively learn data distribution from noisy or incomplete dataset and its internal structure, behavior, and outcomes are easy to understand.
\end{itemize}


\section{Methodology}

\subsection{Generative Model}

\label{sec:Graphical_model}
\begin{wrapfigure}{r}{0.52\textwidth}
\centering
\includegraphics[width=0.48\textwidth]{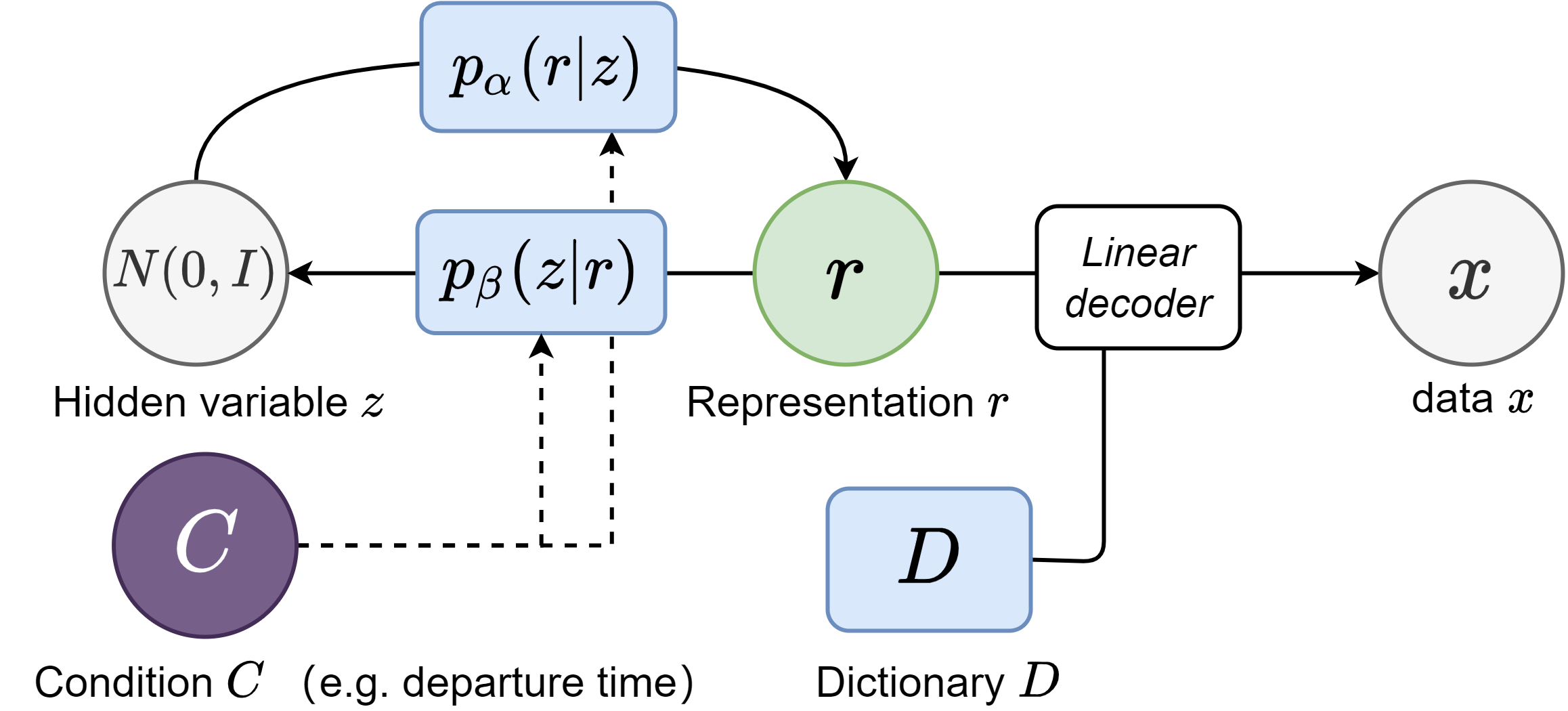}
\caption{\label{fig:pic2} Illustration of the generative framework which describes the generative process of path.}
\end{wrapfigure}

In this work, we formulate the compositional trajectory generation problem using the following Markov chain: $\boldsymbol{z} \rightarrow \boldsymbol{r}  \rightarrow \boldsymbol{x}$. Let $\boldsymbol{x}\in\{0,1\}^{|E|}$ denote the observed trajectory on a road network with edge set $E$, and $\boldsymbol{r}\in\{0,1\}^{n}$ be its sparse binary representation over a pathlet dictionary of size $n$. We introduce a $K$-dimensional latent Gaussian variable $\boldsymbol{z}\in \mathbb{R}^{K}$ that generates the representation $\boldsymbol{r}$.

To enable gradient-based optimization, we adopt a continuous relaxation of the binary trajectory and assume a Gaussian observation model:
\begin{equation}
    \boldsymbol{z} \sim \mathcal{N}(\mathbf{0}, I_K), \qquad
    \boldsymbol{x} \mid \boldsymbol{r} \sim \mathcal{N}(D\boldsymbol{r}, \sigma^2 I_{|E|}),
\end{equation}
where $D\in\{0,1\}^{|E|\times n}$ is the pathlet dictionary matrix and $\sigma^2$ is a fixed noise variance. Although $\boldsymbol{x}$ is represented as a binary vector in $\{0,1\}^{|E|}$, the Gaussian observation model serves as a continuous approximation that facilitates differentiable training.

The dependency between $\boldsymbol{z}$ and $\boldsymbol{r}$ is modeled using a Binary VAE architecture \cite{zhaoVariationalAutoencodersSparse2020}. Given $\boldsymbol{z}$, two neural networks $f_{\alpha^{m}}$ and $f_{\alpha^{p}}$ produce the parameters $\boldsymbol{m}$ and $\boldsymbol{p}$ of a hierarchical Bernoulli distribution:
\begin{equation}
\label{eq:bvae_params}
    \boldsymbol{m} = \exp\!\big(f_{\alpha^{m}}(\boldsymbol{z})\big), \qquad
    \boldsymbol{p} = \operatorname{sigmoid}\!\big(f_{\alpha^{p}}(\boldsymbol{z})\big),
\end{equation}
and each dimension of $\boldsymbol{r}$ is generated as
\begin{equation}
    r_j \mid \boldsymbol{z} \sim \operatorname{Bernoulli}\!\big(1 - p_j^{m_j}\big), \quad j = 1,\dots,n,
\end{equation}
where $p_j$ and $m_j$ denote the $j$-th components of $\boldsymbol{p}$ and $\boldsymbol{m}$, respectively, and $p_j^{m_j}$ denotes element-wise exponentiation.

Under this construction, the joint distribution of all variables factorizes as
\begin{equation}
\label{eq:joint}
    p(\boldsymbol{x},\boldsymbol{r},\boldsymbol{z})
    = p(\boldsymbol{x}\mid\boldsymbol{r};D)\,
      p_{\alpha}(\boldsymbol{r}\mid\boldsymbol{z})\,
      p(\boldsymbol{z}),
\end{equation}
where $p(\boldsymbol{z})=\mathcal{N}(\mathbf{0},I_K)$ and $p_{\alpha}(\boldsymbol{r}\mid\boldsymbol{z})$
is induced by the Bernoulli model above.

\subsection{Variational learning}

From the joint distribution in Eq.~\eqref{eq:joint}, the negative log-density can be decomposed as
\begin{equation}
\label{eq:joint_decomp}
    -\log p(\boldsymbol{x},\boldsymbol{r},\boldsymbol{z})
    = -\log p(\boldsymbol{x}\mid\boldsymbol{r};D)
      -\log p_{\alpha}(\boldsymbol{r}\mid\boldsymbol{z})
      -\log p(\boldsymbol{z}).
\end{equation}
Since $p(\boldsymbol{r},\boldsymbol{z}) = p_{\alpha}(\boldsymbol{r}\mid\boldsymbol{z}) p(\boldsymbol{z})$, the last two terms can be grouped together, yielding
\begin{equation}
    -\log p(\boldsymbol{x},\boldsymbol{r},\boldsymbol{z})
    = -\log p(\boldsymbol{x}\mid\boldsymbol{r};D)
      -\log p(\boldsymbol{r},\boldsymbol{z}).
\end{equation}
This decomposition shows that the overall objective naturally splits into two complementary parts: the first term $-\log p(\boldsymbol{x}\mid\boldsymbol{r};D)$ is captured by the MDL-based dictionary learning objective in Section~4.3, which characterizes the generative process of trajectories conditioned on $\boldsymbol{r}$; the second term $-\log p(\boldsymbol{r},\boldsymbol{z})$ is modeled by the Binary VAE that learns a prior over the representation vectors, as described in this subsection.

Given a collection of representation vectors, our goal is to learn the generative model $p_{\alpha}(\boldsymbol{r}\mid\boldsymbol{z})p(\boldsymbol{z})$ by maximizing the marginal log-likelihood $\sum_i \log p(\boldsymbol{r}^{(i)})$, where
\begin{equation}
    \log p(\boldsymbol{r}) = \log \int p_{\alpha}(\boldsymbol{r}\mid\boldsymbol{z})\,p(\boldsymbol{z})\,d\boldsymbol{z}.
\end{equation}
The integral over $\boldsymbol{z}$ is intractable for a flexible neural decoder, so we resort to variational inference.
We introduce an encoder network $g_{\phi}$ that parameterizes a variational posterior
\begin{equation}
    q_{\phi}(\boldsymbol{z}\mid\boldsymbol{r})
    = \mathcal{N}\big(\boldsymbol{\mu}_{\phi}(\boldsymbol{r}),
      \operatorname{diag}(\boldsymbol{\sigma}_{\phi}^2(\boldsymbol{r}))\big),
\end{equation}
and use the reparameterization trick to draw samples from $q_{\phi}(\boldsymbol{z}\mid\boldsymbol{r})$.

The evidence lower bound (ELBO) for $\log p(\boldsymbol{r})$ is given by
\begin{equation}
    \mathcal{L}_{\text{ELBO}}(\boldsymbol{r})
    = \mathbb{E}_{q_{\phi}\left(\boldsymbol{z} \mid \boldsymbol{r}\right)}
        \left[\log p_{\alpha}\left(\boldsymbol{r} \mid \boldsymbol{z}\right)\right]
      - \operatorname{KL}\!\left[q_{\phi}\left(\boldsymbol{z} \mid \boldsymbol{r}\right) \,\middle\|\, p\left(\boldsymbol{z}\right)\right],
\end{equation}
which satisfies $\mathcal{L}_{\text{ELBO}}(\boldsymbol{r}) \le \log p(\boldsymbol{r})$.
For our Bernoulli decoder, the reconstruction term can be written as
\begin{equation}
   \mathbb{E}_{q_{\phi}\left(\boldsymbol{z} \mid \boldsymbol{r}\right)}
   \left[\log p_{\alpha}\left(\boldsymbol{r} \mid \boldsymbol{z}\right)\right]
   = \sum_{j=1}^{n}
     \mathbb{E}_{q_{\phi}(\boldsymbol{z}\mid\mathbf{r})}
     \left[\boldsymbol{r}_j \log \big(1-\boldsymbol{p}^{\boldsymbol{m}}_j\big)
           + (1 - \boldsymbol{r}_j) \log\big(\boldsymbol{p}^{\boldsymbol{m}}_j\big)\right],
\end{equation}
where $\boldsymbol{m}$ and $\boldsymbol{p}$ are functions of $\boldsymbol{z}$ as defined in Eq.~\eqref{eq:bvae_params}. The second component is the Kullback--Leibler (KL) divergence, which measures the discrepancy between the learned posterior $q_{\phi}\left(\boldsymbol{z} \mid \boldsymbol{r}\right)$ and the prior distribution $p\left(\boldsymbol{z}\right)=\mathcal{N}(\mathbf{0},I_K)$.

In practice, we minimize the negative ELBO as the Binary VAE loss:
\begin{equation}
\label{eq:lvae}
L_{\text{VAE}}
    = -\mathbb{E}_{q_{\phi}\left(\boldsymbol{z} \mid \boldsymbol{r}\right)}
        \left[\log p_{\alpha}\left(\boldsymbol{r} \mid \boldsymbol{z}\right)\right]
      + \operatorname{KL}\!\left[q_{\phi}\left(\boldsymbol{z} \mid \boldsymbol{r}\right) \,\middle\|\, p\left(\boldsymbol{z}\right)\right],
\end{equation}
which encourages the model to accurately reconstruct $\boldsymbol{r}$ while keeping the latent
representation $\boldsymbol{z}$ close to the Gaussian prior.

\subsection{MDL-based dictionary learning}

We use a compositional learning based approach to automatically learn the pathlet dictionary and representation from dataset. Based on the Minimum Description Length (MDL) principle \cite{hansen2001model}, we extend the original reconstruction error term $-\log p(\boldsymbol{x}|\boldsymbol{r}; D)$ to include three components: (1) the reconstruction error, (2) a dictionary complexity regularization term, and (3) a representation sparsity regularization term. The latter two terms are derived from the MDL principle as follows.

In the MDL framework, the total description length for a data set is decomposed into the length for encoding the model $L(H)$ and the length for describing the data given the model $L(D|H)$. Here, the matrix $D$ serves as the model (dictionary), and $R$ is its corresponding sparse data representation. A visualization of the relationship between these matrices $D$, $X$, and $R$ can be found in Figure~\ref{fig:matrices} in Section~\ref{sec:matrix_visualization} of the Appendix.

\textbf{Model Length:}
Each dictionary atom corresponds to a column of $D \in \{0,1\}^{|E| \times n}$, where the binary entries encode the inclusion of graph edges (or grids) in each atom. Since $D$ is binary, storing one atom requires $|E|$ bits. The total storage cost of the effective part of the dictionary (i.e., atoms utilized in the representation) is then: $L(H) = |E| \cdot |\mathcal{A}_{\mathrm{eff}}|$, where $|\mathcal{A}_{\mathrm{eff}}|$ denotes the number of dictionary atoms that are actually used.

The effective dictionary size is determined by examining the usage matrix $R$. An atom $D^{(j)}$ (where $D^{(j)}$ denotes the $j$-th column of matrix $D$) is effective if its corresponding row $R_j$ (where $R_j$ denotes the $j$-th row of matrix $R$) contains any nonzero element: $\mathbf{1}_{\{D^{(j)} \text{ is effective}\}} = \mathbf{1}_{\{\max_i R_{ji} > 0\}}$, where $R_{ji}$ denotes the $(j,i)$-th element of matrix $R$, and $\mathbf{1}_{\{condition\}}$ denotes the indicator function, which equals $1$ if the condition holds and $0$ otherwise. For binary $R$, the effective dictionary size becomes:
\begin{equation}
\label{eq:eff_dict_size}
|\mathcal{A}_{\mathrm{eff}}| = \sum_{j=1}^n \mathbf{1}_{\{\max_i R_{ji} > 0\}} = \sum_{j=1}^n \max_i R_{ji}
\end{equation}


\textbf{Data Representation Length:}
Given a fixed $D$, the representation $R$ is a binary sparse matrix. The length to encode $R$ is determined by the number of nonzero elements, therefore $L(D|H) = \|R\|_0$.
And for binary $R$, the $L_0$ pseudo-norm coincides with the $L_1$ norm.
In summary, the MDL for this dictionary learning setting takes the form
\begin{equation}
\begin{aligned}
\text{MDL} &= L(H) + L(D|H) = |E| \cdot |\mathcal{A}_{\mathrm{eff}}| + \|R\|_1 = |E| \cdot \sum_{j=1}^m \max_i R_{ji} + \|R\|_1
\end{aligned}
\end{equation}

For the reconstruction error term, given our assumption that $\boldsymbol{x} = D\boldsymbol{r} + \epsilon$ where $\epsilon$ represents Gaussian noise, we have: $P(\boldsymbol{x}|\boldsymbol{r}; D) = \mathcal{N}(\boldsymbol{x}; D\boldsymbol{r}, I)$. Maximizing this probability is equivalent to minimizing $\|\boldsymbol{x} - D\boldsymbol{r}\|_2^2$.

Combining the reconstruction error with the MDL-derived regularization terms, the complete compositional learning loss function is defined as:
\begin{equation}
\label{eq:ldict}
L_{\text{dict}} = \|X - DR\|_2^2 + \lambda_1 \sum_{j=1}^m \max_i R_{ji} + \lambda_2 \|R\|_1
\end{equation}
where $\lambda_1$ and $\lambda_2$ are hyperparameters that control the trade-off between reconstruction accuracy and model complexity. The first term ensures faithful reconstruction of the original paths, while the second and third terms encourage dictionary compactness and representation sparsity, respectively.
This loss function $L_{\text{dict}}$ is consistent with the one proposed in \cite{tangExplainableTrajectoryRepresentation2023}. However, our work provides a new interpretation based on the MDL principle, which offers a principled theoretical foundation for understanding why this particular combination of terms leads to effective dictionary learning.

\textbf{Problem formulation:}
Combining both components, our complete objective function is:
\begin{equation}
\min_{\alpha, \beta, D, R} L_{\text{VAE}} + L_{\text{dict}}, \quad \text{s.t.} \quad D_{i,j}\in\{0,1\}, \; R_{i,j}\in\{0,1\}
\end{equation}


\subsection{Training Procedure}

To solve the aforementioned problem, we have designed an end-to-end alteranting optimization algorithm. Since neural networks require differentiable operations for gradient-based optimization, we first relax the integer optimization constraints to interval constraints $[0,1]$, then obtain a decimal solution using projected gradient descent, and finally convert it into a binary solution through probabilistic rounding method. The detailed procedures are provided in Algorithm 1 in the Appendix B.

Our approach essentially involves learning the distribution of a given dataset and then sampling from it to generate new paths. Therefore, an unavoidable issue is that the connectivity of the generated paths is not strictly guaranteed. To address this problem, we design specific post-processing algorithm with detailed procedures provided in Algorithm 2  in the Appendix B.
\subsection{Downstream tasks}
Once the generative model has been obtained, it captures the distribution of the data. Our proposed generative model can be conveniently applied to various downstream tasks, demonstrating its versatility and practical value. We focus on two key applications: conditional path generation and data denoising with noisy paths. 

The conditional generation capability allows users to specify certain conditions or attributes to generate data that meets specific requirements, which can be achieved by replacing the VAE with a CVAE. Figure~\ref{fig:cvae} illustrates the architecture of conditional Pathlet-VAE used in our framework. Additionally, our framework's inherent sparsity and dictionary-based design make it naturally robust to noise, enabling effective data denoising applications. For detailed methodologies, experimental setups, and comprehensive results of these downstream tasks, please refer to Appendix B.

\section{Experiment}
In this section, we conduct extensive experiments to comprehensively evaluate the proposed framework from four aspects: effectiveness, efficiency, robustness, and interpretability, and compare it with previous methods. We first briefly introduce the datasets, experimental setup, and evaluation protocol. After that, we will attempt to answer the following research questions:

\noindent\textbf{RQ1:} Is our proposed generative model capable of learning the true data distribution and generating datasets that closely approximate the true distribution? \noindent\textbf{RQ2:} Can our algorithm be used on noisy or incomplete data? To what extent of noise can it operate normally? \noindent\textbf{RQ3:} Can the model be conveniently applied to different downstream tasks while achieving good performance? \noindent\textbf{RQ4:} What does the representation vector generated signify? Can it be visualized or understood in an intuitive way? \noindent\textbf{RQ5:} How efficient is the proposed method?

\subsection{Experiment setup}
We conduct extensive experiments on multiple real-world taxi trajectory datasets. To validate robustness, we simulate two types of data issues: data incompleteness and measurement noise. For road network-based representation, we randomly replace trajectory nodes with an "unknown" node following a Bernoulli distribution (data incompleteness) with $p_{noise}$ varying across levels. For grid-based representation (measurement noise), we add zero-mean Gaussian perturbations to raw GPS coordinates before gridding, with standard deviation $\sigma$. 
Detailed experimental setup, including dataset descriptions, data preprocessing steps, noise simulation protocols, and baseline model configurations, are provided in Appendix C.

\subsection{Trajectory Generation Results}

\begin{table*}[t]
\centering
\caption{The performance comparison with previous work across varying noise levels. Lower is better. The best
results are highlighted in \textbf{bold}, while the second-best results are \underline{underlined}.}
\small 
\setlength{\tabcolsep}{5pt}
\begin{tabular}{@{}lcccccccccccc@{}}
\toprule
\multicolumn{13}{c}{\cellcolor{gray!30}Road Network-based Representation} \\
\midrule
Dataset                             & \multicolumn{6}{c}{ShenZhen}                                                                                       & \multicolumn{6}{c}{Porto}                                                                     \\ \midrule
\multicolumn{1}{l|}{Noise level}             & 0             & 0.02          & 0.04          & 0.06          & 0.08          & \multicolumn{1}{c|}{0.10}           & 0             & 0.02          & 0.04          & 0.06          & 0.08          & 0.10           \\
\multicolumn{1}{l|}{Noisy data}              & 0.00          & 0.04          & 0.09          & 0.12          & 0.15          & \multicolumn{1}{c|}{0.17}          & 0.00          & 0.05          & 0.10          & 0.14          & 0.18          & 0.22          \\ \midrule
\multicolumn{1}{l|}{MTnet \cite{wangDeepGenerativeModel2022}}                   & 0.08          & 0.20          & 0.23          & 0.24          & 0.25          & \multicolumn{1}{c|}{0.28}          & 0.14          & 0.21          & 0.28          & 0.28          & 0.29          & 0.29          \\
\multicolumn{1}{l|}{GDP \cite{shiGRAPHCONSTRAINEDDIFFUSIONENDTOEND2024}}                     & 0.09          & 0.14          & 0.17          & 0.18          & {\underline{0.18}}    & \multicolumn{1}{c|}{{\underline{0.19}}}    & \textbf{0.07} & 0.17          & {\underline{0.18}}    & {\underline{0.15}}    & {\underline{0.18}}    & {\underline{0.21}}    \\
\multicolumn{1}{l|}{Binary VAE}              & 0.10          & 0.13          & 0.15          & {\underline{0.16}}    & 0.18          & \multicolumn{1}{c|}{0.20}          & 0.13          & 0.17          & 0.19          & 0.22          & 0.27          & 0.27          \\
\multicolumn{1}{l|}{L1B-VAE} & {\underline{0.07}}    & {\underline{0.11}}    & {\underline{0.13}}    & 0.17          & 0.18          & \multicolumn{1}{c|}{0.20}          & {\underline{0.10}}    & {\underline{0.16}}    & 0.19          & 0.23          & 0.26          & 0.27          \\
\multicolumn{1}{l|}{Our method}              & \textbf{0.07} & \textbf{0.07} & \textbf{0.09} & \textbf{0.11} & \textbf{0.13} & \multicolumn{1}{c|}{\textbf{0.15}} & 0.11          & \textbf{0.11} & \textbf{0.13} & \textbf{0.15} & \textbf{0.16} & \textbf{0.18} \\
\midrule
\multicolumn{13}{c}{\cellcolor{gray!30}Grid-based Representation} \\
\midrule
Dataset                             & \multicolumn{6}{c}{ShenZhen}                                                                                       & \multicolumn{6}{c}{Porto}                      \\ \midrule
\multicolumn{1}{l|}{Noise level}    & 0             & 2.5e-4        & 5e-4          & 8e-4          & 1.25e-3       & \multicolumn{1}{c|}{2.5e-3}        & 0    & 2.5e-4 & 5e-4 & 8e-4 & 1.25e-3 & 2.5e-3 \\
\multicolumn{1}{l|}{Noisy data}            & 0.00          & 0.02          & 0.06          & 0.09          & 0.13          & \multicolumn{1}{c|}{0.18}          & 0.00 & 0.03 & 0.09 & 0.13 & 0.17 & 0.20 \\
\midrule
\multicolumn{1}{l|}{Diffusion}      & 0.08          & 0.10          & 0.11          & 0.14          & 0.17          & \multicolumn{1}{c|}{0.19}          & 0.13 & 0.16 & 0.17 & 0.20 & 0.21 & 0.22 \\
\multicolumn{1}{l|}{diff-lstm \cite{zhuDiffTrajGeneratingGPS2023}}      & 0.04          & 0.05          & 0.09          & 0.11          & \underline{0.14}          & \multicolumn{1}{c|}{0.18}          & 0.12 & 0.15 & 0.16 & 0.17 & 0.18 & 0.22 \\
\multicolumn{1}{l|}{diffwave \cite{kong2021diffwave}}       & 0.06          & 0.08          & 0.09          & 0.13          & 0.17          & \multicolumn{1}{c|}{0.19}          & 0.12 & 0.13 & 0.15 & 0.17 & 0.19 & 0.21 \\
\multicolumn{1}{l|}{diff-traj\cite{zhuDiffTrajGeneratingGPS2023}}      & \underline{0.03}          &\underline{ 0.05  }        & \underline{0.08}          & \underline{0.11}          & 0.15          & \multicolumn{1}{c|}{\underline{0.17}}          & \underline{0.08} & \underline{0.10} & \underline{0.13} & \underline{0.15} & \underline{0.18} & \underline{0.21} \\ \midrule
\multicolumn{1}{l|}{Our method}           & \textbf{0.01} & \textbf{0.02} & \textbf{0.06} & \textbf{0.10} & \textbf{0.14} & \multicolumn{1}{c|}{\textbf{0.14}} & \textbf{0.02} & \textbf{0.04} & \textbf{0.10} & \textbf{0.14} & \textbf{0.17} & \textbf{0.20} \\ \bottomrule
\end{tabular}
\end{table*}

\textbf{Evaluation Protocol and Baselines:} We use Jensen-Shannon Divergence (JSD) to quantitatively measure the distribution distance between generated and real trajectory datasets. We compare our method against multiple state-of-the-art baselines across both road network-based and grid-based representations. 

\textbf{Numerical Results:} Table 1 presents a comprehensive numerical comparison of our approach with other methods under different noise levels across both road network-based and grid-based representations. Results show that in the absence of noise, the datasets generated by various methods exhibit similar performance. As the noise level increases, all methods produce datasets that deviate further from the true distribution. However, our method demonstrates superior robustness across both representation types.

For road network-based representation, traditional methods like MTnet and GDP show significant performance degradation as noise increases. On the Shenzhen dataset, when the noise level exceeds 0.04, our method generates datasets closer to the true distribution than even the noise-affected original data. 
For grid-based representation, the performance improvements are even more pronounced. Diffusion-based methods  show substantial degradation under noise. Notably, our method achieves the best performance across all noise levels on both datasets, with particularly significant improvements on the Porto dataset where the performance gap widens as noise increases. Additionally, experiments regarding data recovery from incomplete inputs are provided in Figure \ref{fig:data_denoising}  (Appendix E). 
\emph{These comprehensive experiments across both representation types demonstrate that our method can effectively learn robust data distributions and significantly outperforms existing approaches under noisy conditions.} (For \textbf{RQ1-2} )

\begin{table}[t]
\centering
\begin{minipage}{0.48\linewidth}
\vspace{-1.2em}
\centering
\caption{Performance comparison of data efficiency.}
\label{tab:data-efficiency}
\setlength{\tabcolsep}{4pt}
\begin{tabular}{@{}lcccccc@{}}
\toprule
& \multicolumn{3}{c}{ShenZhen} & \multicolumn{3}{c}{Porto} \\
\cmidrule(lr){2-4} \cmidrule(lr){5-7}
Method & 5\% & 20\% & 80\% & 5\% & 20\% & 80\% \\
\midrule
Diffusion & 0.19 & 0.07 & 0.05 & 0.25 & 0.16 & 0.06 \\
diff-lstm & {\underline{0.16}} & 0.12 & 0.13 & {\underline{0.17}} & 0.11 & 0.05 \\
diffwave & 0.18 & 0.07 & {\underline{0.04}} & 0.18 & 0.08 & 0.05 \\
diff-traj & 0.18 & {\underline{0.06}} & 0.04 & 0.17 & {\underline{0.07}} & {\underline{0.05}} \\
\midrule
Ours & \textbf{0.01} & \textbf{0.01} & \textbf{0.01} & \textbf{0.06} & \textbf{0.04} & \textbf{0.04} \\
\bottomrule
\end{tabular}
\end{minipage}
\begin{minipage}{0.51\textwidth}
\centering
\caption{Next-edge prediction performance on the Shenzhen and Porto vehicle trajectory datasets.
We report the average cross-entropy loss (lower is better) over all time steps in the test set for each city.}
\label{tab:next_edge_prediction}
\small
\setlength{\tabcolsep}{3pt}
\begin{tabular}{lccc}
\toprule
Method              & Shenzhen $\downarrow$ & Porto $\downarrow$ & Avg $\downarrow$ \\
\midrule
Uniform (outgoing)  & 1.51 & 1.53 & 1.52 \\
First-order Markov  & 1.11 & 1.20 & 1.15 \\
LSTM-Edge           & 0.94 & 0.98 & 0.96 \\
MTnet \cite{wangDeepGenerativeModel2022} & 0.90 & 0.95 & 0.93 \\
\midrule
Our method          & \textbf{0.81} & \textbf{0.88} & \textbf{0.84} \\
\bottomrule
\end{tabular}
\end{minipage}
\vspace{-1.2em}
\end{table}

Table 2 demonstrates the superior data efficiency of our approach. When training data is limited to 5\% of the full dataset, our method achieves JSD values of 0.01 on Shenzhen and 0.06 on Porto, significantly outperforming baseline methods. Our method maintains consistent performance across all data sizes (5\%-80\%), while baseline methods exhibit substantial performance fluctuations. This data efficiency stems from the dictionary learning framework's ability to capture essential mobility patterns with limited samples, making it  valuable for real-world applications where trajectory data collection is expensive or privacy-constrained.

\subsection{Application in trajectory prediction}

\noindent\textbf{Problem setup (next-edge prediction).}
Given a trajectory on the road network $\boldsymbol{t} = (e_1, e_2, \dots, e_L)$, we construct supervised samples from each prefix: at step $t$, the input consists of the prefix $(e_1,\dots,e_t)$ and the target label is the next edge $y_t = e_{t+1}$. 
Let $\mathcal{E}$ denote the set of edges in the road network and $p_{\theta}(e \mid e_{1:t})$ be the model-predicted probability of choosing edge $e$ as the next step.
We evaluate models by the average cross-entropy loss over all time steps in the test set: 
$
\mathcal{L}_{\text{CE}}
= - \frac{1}{N}\sum_{i=1}^N \log p_{\theta}\big(y_i \mid \mathrm{prefix}_i\big)
$, 
where $y_i \in \mathcal{E}$ is the ground-truth next edge for the $i$-th sample and $N$ is the total number of prediction instances.

\noindent\textbf{Baselines.}
We compare our method with several lightweight next-edge predictors, including a uniform outgoing-edge model, a first-order Markov model, and an LSTM-based sequence model. 
Our model uses the learned pathlet-based representation to produce next-edge probabilities under the same setting. 

Table~\ref{tab:next_edge_prediction} shows that while probabilistic baselines and the LSTM-Edge model already capture useful local transition patterns, our method consistently achieves the lowest cross-entropy on both Shenzhen and Porto. 
This indicates that the pathlet-based representation learned by our generative model provides more informative features for downstream prediction.
Appendix~D further presents a conditional trajectory prediction case study and data denoising results, where our method achieves up to 68\% noise reduction on the Porto dataset.
\emph{These downstream experiments demonstrate the utility of our method}.  (For \textbf{RQ3} )


\subsection{Interpretability}
\begin{wrapfigure}{r}{0.42\textwidth}
\centering
\includegraphics[width=0.40\textwidth]{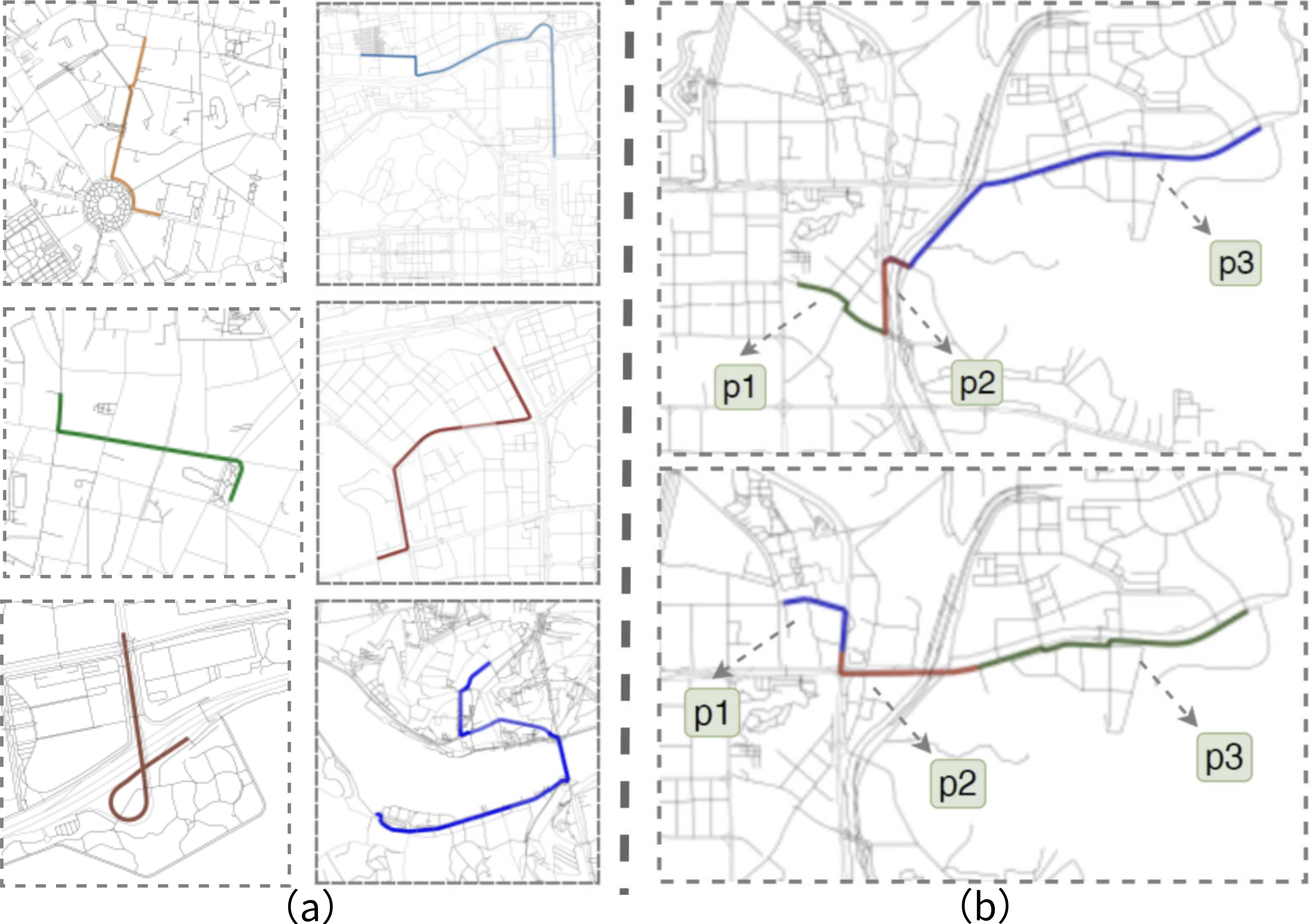}
\caption{Visualization of pathlets and trajectory decomposition.}
\label{fig:learning_process}
\end{wrapfigure}
Besides robustness, incorporating pathlet dictionary learning into the architecture of the generative model offers the additional benefit of enhanced interpretability, which is crucial for trust and usability. 
Figure \ref{fig:learning_process}(a) illustrates the dictionary elements learned by the model, showing meaningful mobility patterns that capture commonly used routes. 
For example, the third row shows a pathlet in Porto, which represents a route crossing from the north bank to the south bank of the Douro River via the Dom Luís I Bridge. This is an important traffic route. 
Figure \ref{fig:learning_process}(b) demonstrates how trajectory samples are concatenated and generated using elements from the dictionary. \emph{The visualization of critical parameters and outcomes within the model assists us in gaining a deeper understanding of the model's behavior.} (For \textbf{RQ4})

\subsection{Efficiency Analysis} 
\begin{wraptable}{r}{0.42\textwidth}
\renewcommand{\arraystretch}{1.2}
\centering
\caption{The efficiency comparison of different methods.}
\label{tab:efficiency-comparison}
\setlength{\tabcolsep}{3pt}
\small
\begin{tabular}{ccccc}
\hline
Method & \begin{tabular}[c]{@{}c@{}}Diction-\\ ary Size\end{tabular} & \begin{tabular}[c]{@{}c@{}}Represent-\\ ation Cost\end{tabular} & \begin{tabular}[c]{@{}c@{}}GPU \\ Memory\end{tabular} & \begin{tabular}[c]{@{}c@{}}Running \\ Time\end{tabular} \\ \midrule
DP & 2.16k & 2.13 & - & 2.2h \\
GD & 1.67k & 2.05 & 30.1G & 0.85h \\
Ours & 1.03k & 1.96 & 13.1G & 0.3h \\ \hline
\end{tabular}
\end{wraptable} 
One major challenge in obtaining a pathlet dictionary from trajectory data is that the space of candidate dictionary elements grows exponentially with the scale of the road network, suffering from the curse of dimensionality. 
\cite{tangExplainableTrajectoryRepresentation2023} use pre-filtering and hierarchical learning strategies to keep the computational cost within an acceptable range, albeit by sacrificing some performance. 
\cite{chenPathletLearningCompressing2013b} employed a dynamic programming approach, which could not leverage the computational acceleration benefits of modern GPUs. (The two aforementioned works are respectively referred to as GD and DP in Table~\ref{tab:efficiency-comparison}.) 
In contrast to prior research, our work is the first to adopt a learning-based rather than a selection-based strategy, which provides significant efficiency advantages to our method.
\emph{It is evident that our scheme not only surpasses the previous ones in terms of dictionary size and expression cost, but also reduces training time by 64.8\% and 86.4\%, respectively, and cuts down GPU memory usage by 56.5\%.} (For \textbf{RQ5})


\section{Conclusion}
In this paper, we introduce a novel framework that integrates VAE and sparse pathlet dictionary learning for trajectory generation. Through experiments on multiple datasets, we found that the learned data distribution under various levels of noise is closer to the real distribution compared to benchmark models. Furthermore, we investigated its performance on downstream tasks and visualized the dictionary learning process, demonstrating that our framework offers more interpretability compared to purely deep neural networks. 

In future work, we will delve deeper into the theoretical principles and guarantees of its robustness, as well as explore its application in a wider range of scenarios.

\newpage
\bibliography{reference}


\appendix
\section{Appendix}
This appendix provides comprehensive supplementary material for the Pathlet Variational Auto-Encoder framework. We present: (1) method details including conditional path generation with CVAE architecture, data denoising capabilities, and a greedy post-processing algorithm for trajectory connectivity; (2) detailed experimental setups covering two datasets (Shenzhen, Porto) with preprocessing procedures, noise simulation protocols, JSD evaluation metrics, and five baseline comparisons; (3) additional experimental results demonstrating data denoising performance; and (4) complete mathematical notation table. These materials provide essential technical details supporting the effectiveness and interpretability of our approach.


%

\section{Matrix Visualization}
\label{sec:matrix_visualization}

\begin{figure}[h]
\centering
\includegraphics[width=0.75\textwidth]{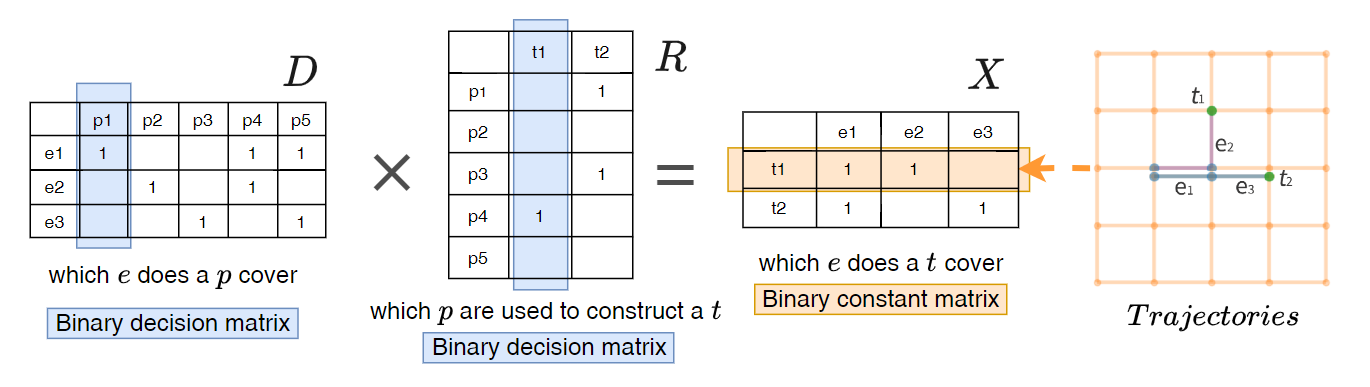}
\caption{\label{fig:matrices} Illustration of the matrices $D,X,R$ (visualized in edge form): $X$ is the trajectory matrix generated from the dataset; $D$ refers to the pathlet dictionary matrix; $R$ is the representation matrix where each column corresponds to a representation vector.}
\end{figure}

\section{Method Details}

\subsection{Conditional Path Generation}
\label{sec:Conditional path generation}
Compared to unconditional generation, conditional generation allows the specification of certain conditions or attributes, thereby generating data that meets specific requirements. This control capability is particularly useful when data with specific characteristics is needed.  
This capability can be achieved in our framework with a simple modification: replacing the VAE with a CVAE. The figure below illustrates a specific scenario where, for example, the first half and departure time of the path are provided as conditions along with the input, and the model generates the complete path. For the encoding method, we map the departure time to 24 hourly intervals and use one-hot encoding. In this case, the framework can perform the path prediction task.
\begin{figure}[h]
\centering
\includegraphics[width=0.76\textwidth]{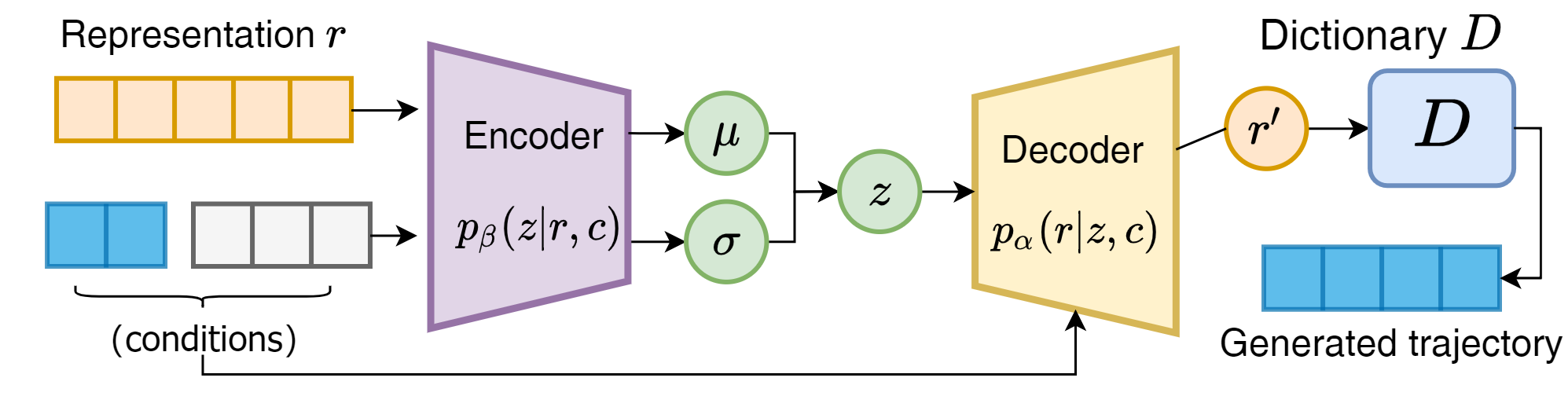}
\caption{\label{fig:cvae} Architecture of Conditional Pathlet-VAE.}
\end{figure}

\subsection{Generation with Noisy Paths}
As mentioned earlier, our framework incorporates the concept of pathlet dictionary and the sparse combination of elements within the dictionary for data generation. This design not only provides interpretability but also introduces sparsity, making the model inherently robust to noise and suitable for data denoising tasks. Specifically, after completing the model training, the dictionary $D$ is fixed as one of the model parameters. For a new noisy data $\boldsymbol{x}$, denoising consists of two steps:

(1) Map $\boldsymbol{x}$ to a new representation space using $D$. To be specific, representation vector $\boldsymbol{r}$ is obtained by solving:
\begin{equation}
    \min\limits_{{\boldsymbol{r}_{i} \in \{0,1\}}} \ \ \lambda_2||\boldsymbol{r}||_1+\lambda_3 ||\boldsymbol{x} - D\boldsymbol{r}||_1
\end{equation}

This problem can be viewed as a simplified version of the original problem because the dictionary is fixed at this moment. We solve it using the same strategy described before: first get the optimal fractional solution $\boldsymbol{r}^*$ using gradient descent and then round it to get the final binary solution $\boldsymbol{r}^-$.

(2) Once we get $\boldsymbol{r}^*$, then the denoised data $\hat{x}$ can be obtained by multiplying $D$ and $\boldsymbol{r}^-$.

\subsection{Train Algorithm}
Here we detail the two-step algorithm outlined in Section 4.4 (Training Procedure). We employ a relax-and-round approach: the binary constraints are relaxed to the continuous domain [0,1] for gradient-based optimization via projected gradient descent, followed by probabilistic rounding to recover the discrete solution.
\begin{algorithm}[h]
    \caption{Training Procedure}  
    \label{alg:algorithm}
    \textbf{Input}:  \ $X$: path matrix;
           \ $\epsilon,\theta$: hyperparameters;
           \ $L$: loss function;
           \ $\delta$: learning rate;\\
    \textbf{Output}:  Solution $R^{r}$ and $D^{r}$
    \begin{algorithmic}[1] 
    \STATE\# Step1, we compute the fractional solution $R^*$ and $D^*$ using gradient descend.
    \STATE Initial $R_0=I$; initial $D_0=X$, initial $\Delta=+\infty$  
    \STATE Randomly initialize the parameters $\alpha,\beta$ of the VAE model.
    \WHILE{$\Delta<\epsilon$}
    \STATE Compute loss: $L_k = L(\alpha_k,\beta_k,R_k,D_k)$ by Eq 15
    \STATE Compute gradient  and update the parameters 
    \begin{align*}  
    R_{k+1} &= R_k-\delta \nabla_{ R_k} L_k \quad D_{k+1} = D_k-\delta \nabla_{ D_k} L_k \\
    \alpha_{k+1} &= \alpha_k-\delta \nabla_{ \alpha_k} L_k \quad \  \beta_{k+1} = \beta_k-\delta \nabla_{ \beta_k} L_k  
    \end{align*}  
          \STATE Clip the result to ensure that every element in  $ R_k$ and $ D_k$  has a value between 0 and 1
        \STATE $\Delta=|L_{k}-L_{k-1}|$
    \ENDWHILE
        \STATE \noindent \#Step2, round solution $R^{r}$ and $D^{r}$ based on $R^*,D^*$.
         \STATE Sample $R$ with $P(R_{i,j}=1)=min(1,\theta R^*_{i,j})$
         \STATE Sample $D$ with $P(D_{i,j}=1)=min(1,\theta D^*_{i,j})$ 
    \end{algorithmic}
    \end{algorithm}
    

\subsection{Post-processing Algorithm}
Our approach essentially involves learning the distribution of a given dataset and then sampling from it to generate new paths. Therefore, an unavoidable issue is that the connectivity of the generated paths is not strictly guaranteed. To address this problem, we propose a greedy algorithm that uses road network information to post-process the generated paths to ensure connectivity.

The detailed steps are explained in Algorithm 2: Given a network $G$ and a generated path on it, assume it is not a completely connected path but a set of separate segments. Our goal is to complete these segments into a full path. 
The basic idea of the algorithm is: For a set of separate segments, each time find the two segments with the smallest connection cost (The connection cost is defined as the length of the shortest path in the  $G$ that connects these two segments), merge them into a new segment, and repeat this process until a complete path is obtained.

\begin{algorithm}[h]  

\caption{Post-processing Procedure}  
\begin{algorithmic}[1]  
\REQUIRE List of path fragments $\texttt{segments}$; Network  $G$; Cost function $\texttt{cost(path)}$  
\ENSURE Merged path $\texttt{merged\_path}$  


\WHILE{Number of fragments in \texttt{segments} $>$ 1}  

    \STATE Initialize minimum cost \texttt{min\_cost} $\leftarrow \infty$  
    \STATE Initialize best fragment pair \texttt{best\_pair} $\leftarrow$ \texttt{None}  

    \FOR{Each pair of fragments $(s_i, s_j)$ in \texttt{segments}, where $i \neq j$}  

        \STATE Compute the minimum cost to connect $s_i$ and $s_j$:  
        \STATE Use the shortest path algorithm to find the shortest path between the start or end points of $s_i$ and $s_j$  
        \STATE Obtain the minimum cost \texttt{cost}  

        \IF{\texttt{cost} $<$ \texttt{min\_cost}}  
            \STATE Update \texttt{min\_cost} $\leftarrow$ \texttt{cost}  
            \STATE Update \texttt{best\_pair} $\leftarrow$ $(s_i, s_j)$  
            \STATE Record the shortest path 
        \ENDIF  

    \ENDFOR

    \STATE Merge \texttt{best\_pair} using shortest path 
    \STATE  Add \texttt{new\_segment} to \texttt{segments}  
    \STATE  Remove $s_i$ and $s_j$ from \texttt{segments}

\ENDWHILE  

\RETURN $\texttt{merged\_path}$  

\end{algorithmic}  
\end{algorithm}





    
        




\section{Detailed Experimental Setup}

\subsection{Datasets}

\noindent\textbf{Shenzhen.} \cite{zhangUrbanCPSCyberphysicalSystem2015} released this dataset containing approximately 510k dense trajectories generated by ~14k taxi cabs in Shenzhen, China, which can be downloaded at \cite{HttpsPeopleCs2022}.

\noindent\textbf{Porto.} This dataset describes trajectories performed by 442 taxis running in the city of Porto, Portugal \cite{moreira-matiasPredictingTaxiPassenger2013}. Each taxi reports its location every 15s. This dataset is used for the Trajectory Prediction Challenge@ ECML/PKDD 2015.


\subsection{Data Preprocessing.} For these two datasets, we remove trajectories with less than 20 GPS sample points and use the method proposed in \cite{meertHMMNonemittingStates2018} to convert trajectory to a series of edges on roadmap. 


\subsection{Data Corruption Simulation.} To validate the robustness of our method, we simulate two types of data issues that cover both representation paradigms: data incompleteness (road network-based) and measurement noise (grid-based). For data incompleteness, we add an "unknown" node connected to all nodes; for each trajectory $x$, each node is randomly replaced by the "unknown" node according to a Bernoulli distribution with parameter $p_{noise}$. By adjusting $p_{noise}$, we obtain datasets at varying missing levels. For measurement noise, we perturb raw GPS coordinates prior to gridding:
\[
 (\mathrm{lon}', \mathrm{lat}') = (\mathrm{lon} + \epsilon_{\mathrm{lon}},\, \mathrm{lat} + \epsilon_{\mathrm{lat}}),\quad \epsilon_{\mathrm{lon}}, \epsilon_{\mathrm{lat}} \sim \mathcal{N}(0, \sigma^2).
\]

\subsection{Evaluation Protocol}

To evaluate whether our framework can effectively learn the data distribution, we use the JSD score to quantitatively measure the distribution distance between the generated trajectory dataset and the real dataset in our experimental section. Specifically, we adopt the same definition as in \cite{wangDeepGenerativeModel2022}:
\[
\begin{array}{r}
\text { JSD }\left(x_{0}\right) = \frac{1}{2}\left[\mathbb{E}_{P_{d}\left(e \in x_{1: L} \mid x_{0}\right)} \log \frac{P_{d}\left(e \in x_{1: L} \mid x_{0}\right)}{A}+\right.\left.\mathbb{E}_{\mathcal{G}\left(e \in \hat{x}_{1: L} \mid x_{0}\right)} \log \frac{\mathcal{G}\left(e \in \hat{x}_{1: L} \mid x_{0}\right)}{A}\right]
\end{array}
\]

Here, $\mathcal{G}$ represents the real trajectory distribution, and $P_{d}\left(e \in x_{1: L} \mid x_{0}\right)$ is the conditioned probability that $e$ is a part of the trajectory given $x_0$ as origin edge. $A$ is the average of these two distributions. A smaller JSD value indicates better trajectory modeling performance because the JSD quantifies the distance between distribution.

\subsection{Baselines}

\noindent\textbf{1) GDP} is a diffusion-based model for path planning and generation that learns path patterns while incorporating road network constraints, which can be considered the state-of-the-art method in road network trajectory generation \cite{shiGRAPHCONSTRAINEDDIFFUSIONENDTOEND2024}.

\noindent\textbf{2) MTNet} is a deep generative model which employs a knowledge-based meta-learning module to learn generalized distribution patterns from skewed trajectory data \cite{wangDeepGenerativeModel2022}.

\noindent\textbf{3) Binary VAE.} This basic binary version VAE is identical to the VAE component in our framework to ensure a fair comparison. It directly learns the distribution of vectorized trajectory data and generates datasets accordingly.

\noindent\textbf{4) L1B-VAE.} This is an improved algorithm based on the original Binary VAE by adding L1 regularization as a penalty term.

\noindent\textbf{5) DiffWave.} This is a diffusion model based on the WaveNet architecture \cite{kong2021diffwave}. It employs a novel denoising diffusion process that iteratively refines random noise into coherent patterns, effectively modeling complex data distributions. This method is commonly used as a baseline reference in previous studies within the trajectory generation field.

\noindent\textbf{6) DiffTraj.} This is an advanced trajectory generation method based on a diffusion probabilistic model \cite{zhuDiffTrajGeneratingGPS2023}, designed to effectively simulate and generate high-quality GPS trajectories. Its innovation lies in the adoption of a novel spatial-temporal diffusion process, which gradually transforms random noise into coherent trajectory patterns, thereby achieving precise modeling of complex data distributions.

\subsection{Implementation Details}

\noindent\textbf{Experimental Configuration.} We randomly sampled 30\% trajectories as our test dataset, and use the rest 70\% as the training dataset. Our method is implemented in Python and trained using an Nvidia A40 GPU. All experiments are run on the Ubuntu 20.04 operating system with an Intel Xeon Gold 6330 CPU.

\noindent\textbf{Hyperparameter Settings.} The dictionary size is set to 1000 for both datasets. 
The regularization parameter $\lambda$ is determined using the theoretical optimal hyperparameter selection method proposed in \cite{Zhao2025AUP}. 
The VAE latent dimension is set to 64. Training is performed for 200 epochs with a batch size of 32 using the Adam optimizer with learning rate 0.001.

\section{Additional Experimental Results}

\subsection{Data Denoising Results}
\begin{figure}[h]
\centering
\includegraphics[width=0.82\textwidth]{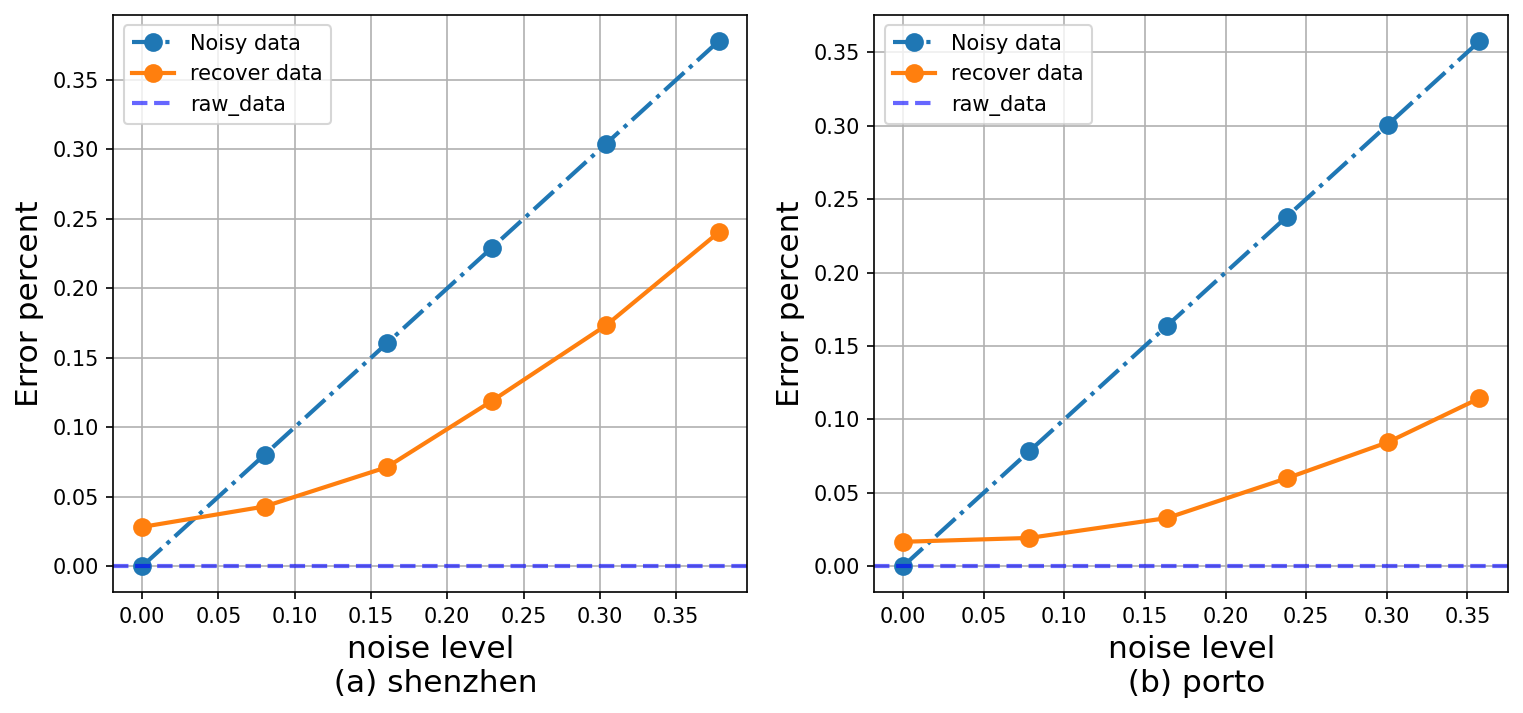}
\caption{\label{fig:data_denoising}Recovery performance under different noise levels.}
\end{figure}

This section demonstrates the results of data denoising using the method described above. Here, the true dataset is denoted as $X$, and the noise-contaminated dataset is denoted as $\tilde{X}$. 
For each trajectory $x$, each node is randomly replaced by the "unknown" node according to a Bernoulli distribution with parameter $p_{noise}$. By adjusting $p_{noise}$, we obtain datasets at varying missing levels.
The model receives $X$ as input and outputs the denoised result $\hat{X}$. The error ratio is used as the evaluation criterion, which is calculated as the difference between $X$ and $\hat{X}$ divided by the L1 norm of $X$. From Figure \ref{fig:data_denoising}, we can see that as the noise increases, the error rate of the restored data is significantly lower than that of the noisy dataset, indicating that our method effectively removes noise. Especially on the Porto dataset, up to about 68\% of the noise can be removed.

\subsection{Case Study: Conditional Trajectory Prediction}

\begin{figure}[h]
    \centering
    \includegraphics[width=0.75\textwidth]{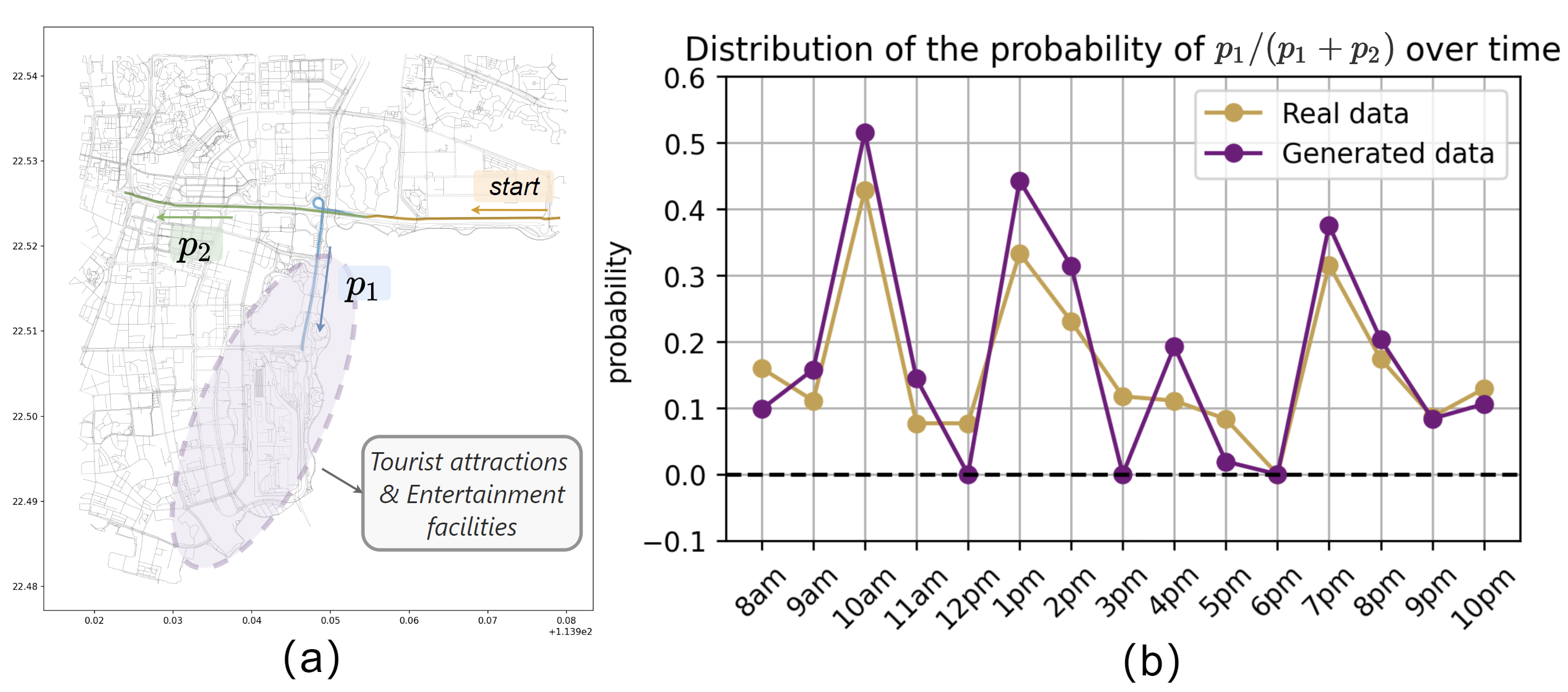}
    \caption{\label{fig:conditional_generation} (a) Visualization of conditionally generated trajectories. (b) Probability distribution of path selection over time.}
\end{figure}

Figure \ref{fig:conditional_generation} presents a case study on the Shenzhen dataset: the first half of the path lies on one of the city's main east--west thoroughfares. 
In the southern region, tourist attractions, dining, and entertainment facilities are densely concentrated, leading to two main trajectory possibilities: going straight or turning left. 

Figure \ref{fig:conditional_generation}(b) illustrates how behavioral patterns on this road segment vary over time: during the morning rush hour, lunchtime, and dinner, there is a significant increase in the probability of people turning left towards the southern area, while at other times, east--west commuting is predominant. 
These results show that our proposed model has effectively learned the underlying spatiotemporal distribution patterns and can capture multi-modal route choices under different temporal conditions.


\begin{table}[h]
\centering
\caption{Summary of Mathematical Notation}
\label{tab:notation}
\footnotesize
\setlength{\tabcolsep}{3pt}
\begin{minipage}[t]{0.49\textwidth}
\vspace{0pt}
\begin{tabular}{@{\hspace{2pt}}p{0.25\textwidth}@{\hspace{4pt}}p{0.65\textwidth}@{\hspace{2pt}}}
\toprule
\multicolumn{2}{c}{\cellcolor{gray!20}\textbf{Basic Variables}} \\
\midrule
$X$ & Trajectory dataset \\
$Y$ & Generated trajectory data \\
$\boldsymbol{t}$ & Individual trajectory \\
$\boldsymbol{x}$ & Binary vector representation of trajectory \\
$\boldsymbol{r}$ & Binary representation vector using pathlet dictionary \\
$\boldsymbol{z}$ & Latent Gaussian variable \\
$D$ & Pathlet dictionary matrix \\
$R$ & Representation matrix \\
$M$ & Generative model \\
\midrule
\multicolumn{2}{c}{\cellcolor{gray!20}\textbf{Graph \& Network}} \\
\midrule
$G$ & Directed graph (road network) \\
$E$ & Set of edges in graph \\
$V$ & Set of vertices in graph \\
$|E|$ & Number of spatial units (edges or grid cells) \\
$e$ & Edge in road network \\
$c$ & Grid cell \\
$p$ & Pathlet (trajectory segment) \\
\midrule
\multicolumn{2}{c}{\cellcolor{gray!20}\textbf{Model Parameters}} \\
\midrule
$n$ & Dictionary size (number of pathlets) \\
$K$ & Dimension of latent variable $\boldsymbol{z}$ \\
$\alpha, \beta$ & Neural network parameters \\
$\lambda_1, \lambda_2$ & Regularization hyperparameters \\
$\theta$ & Rounding parameter \\
$\epsilon$ & Gaussian noise term \\
$\sigma^2$ & Variance parameter \\
$p_{noise}$ & Noise probability parameter \\
\bottomrule
\end{tabular}
\end{minipage}
\hfill
\begin{minipage}[t]{0.49\textwidth}
\vspace{0pt}
\begin{tabular}{@{\hspace{2pt}}p{0.28\textwidth}@{\hspace{4pt}}p{0.62\textwidth}@{\hspace{2pt}}}
\toprule
\multicolumn{2}{c}{\cellcolor{gray!20}\textbf{Matrix \& Vector Operations}} \\
\midrule
$D^{(j)}$ & $j$-th column of matrix $D$ \\
$R_j$ & $j$-th row of matrix $R$ \\
$R_{ji}$ & $(j,i)$-th element of matrix $R$ \\
$\boldsymbol{x}_i$ & $i$-th element of vector $\boldsymbol{x}$ \\
$\|\cdot\|_0$ & $L_0$ pseudo-norm (number of non-zero elements) \\
$\|\cdot\|_1$ & $L_1$ norm \\
$\|\cdot\|_2$ & $L_2$ norm (Euclidean norm) \\
\midrule
\multicolumn{2}{c}{\cellcolor{gray!20}\textbf{Functions \& Operators}} \\
\midrule
$\mathbf{1}_{\{\cdot\}}$ & Indicator function \\
$\max_i$ & Maximum over index $i$ \\
$\sum_{j=1}^n$ & Summation from $j=1$ to $n$ \\
$\exp(\cdot)$ & Exponential function \\
$\log(\cdot)$ & Logarithm function \\
$\text{sigmoid}(\cdot)$ & Sigmoid activation function \\
$f_{\alpha^m}(\cdot)$ & Neural network function for parameter $m$ \\
$f_{\alpha^p}(\cdot)$ & Neural network function for parameter $p$ \\
\midrule
\multicolumn{2}{c}{\cellcolor{gray!20}\textbf{Loss Functions \& Objectives}} \\
\midrule
$L_{\text{VAE}}$ & Variational autoencoder loss \\
$L_{\text{dict}}$ & Dictionary learning loss \\
$L(H)$ & Model description length \\
$L(D|H)$ & Data description length given model \\
$\text{MDL}$ & Minimum description length \\
$|\mathcal{A}_{\text{eff}}|$ & Number of effective dictionary atoms \\
\bottomrule
\end{tabular}
\end{minipage}
\end{table}

\end{document}